\def\year{2019}\relax
\documentclass[letterpaper]{article} 
\usepackage{aaai19}  
\usepackage{times}  
\usepackage{helvet}  
\usepackage{courier}  
\usepackage{url}  
\usepackage{graphicx}  
\usepackage{algorithmic}
\usepackage[ruled]{algorithm2e}
\usepackage{booktabs}
\usepackage{amsmath}
\usepackage{amsthm}
\usepackage{subfigure}
\usepackage{amsfonts}
\usepackage{microtype}
\usepackage{graphicx}
\usepackage{booktabs}
\usepackage{multirow}
\usepackage{bm}
\usepackage{listings}

\begin{document}
%

\title{The Adversarial Attack and Detection under the Fisher Information Metric}
\author{
  Chenxiao Zhao\thanks{Email: \texttt{51174506043@stu.ecnu.edu.cn}}$^1$\ \ P. Thomas Fletcher$^2$\ \ Mixue Yu$^1$\ \ Yaxin Peng$^{3,4}$\ \ Guixu Zhang$^1$
\ \ Chaomin Shen\thanks{Corresponding author. Email: \texttt{cmshen@cs.ecnu.edu.cn}}$^{1,4}$ \\
$^1$Department of Computer Science, East China Normal University, Shanghai, China\\
$^2$Department of Electrical and Computer Engineering, and Department of Computer Science,
University of Virginia, Virginia, USA\\
$^3$Department of Mathematics, Shanghai University, Shanghai, China\\
$^4$Westlake Institute for Brain-Like Science and Technology, Zhejiang, China\\
}
\maketitle

\begin{abstract}
  Many deep learning models are vulnerable to the adversarial attack, i.e.,
  imperceptible but intentionally-designed perturbations to the input can cause incorrect output of the networks.
  In this paper, using information geometry, we provide a reasonable explanation for the vulnerability of deep learning models.
  By considering the data space as a non-linear space with the Fisher information metric induced from a neural network,
  we first propose an adversarial attack algorithm termed one-step spectral attack (OSSA).
  The method is described by a constrained quadratic form of the Fisher information matrix,
  where the optimal adversarial perturbation is given by the first eigenvector,
  and the vulnerability is reflected by the eigenvalues.
  The larger an eigenvalue is, the more vulnerable the model is to be attacked by the corresponding eigenvector.
  Taking advantage of the property, we also propose an adversarial detection method with the eigenvalues serving as characteristics.
  Both our attack and detection algorithms are numerically optimized to work efficiently on large datasets.
  Our evaluations show superior performance compared with other methods,
  implying that the Fisher information is a promising approach to investigate the adversarial attacks and defenses.
\end{abstract}

\section{Introduction}
Deep learning models have achieved substantial achievements on various of computer vision tasks.
Recent studies suggest that, however, even though a well-trained neural network generalizes well on the test set,
it is still vulnerable to adversarial attacks \cite{szegedy2013intriguing}.
For image classification tasks, the perturbations applied to the images can be imperceptible for human perception,
meanwhile misclassified by networks with a high rate.
Moreover, empirical evidence has shown that the adversarial examples have the ability to transfer among different deep learning models.
The adversarial examples generated from one model can often fool other models which have totally different structure and parameters
\cite{papernot2016transferability},
thus making the malicious black-box attack possible.
Many deep learning applications, e.g. the automated vehicles and face authentication system,
have low error-tolerance rate and are sensitive to the attacks.
The existence of adversarial examples has raised severe challenges for deep learning models in security-critical computer vision applications.

Understanding the mechanism of adversarial examples is a fundamental problem for defending against the attacks.
Many explanations have been proposed from different facets.
\cite{szegedy2013intriguing} first observes the existence of adversarial examples,
and suggests it is due to the excessive non-linearity of the neural networks.
On the contrary, \cite{Goodfellow2014Explaining} suggests that the vulnerability results from the models being too linear.
Despite its contradiction to the general impression, the explanation is supported by numbers of experimental results \cite{Krotov2017Dense,tabacof2016exploring,Tanay2016A,tramer2017space}.
On the other hand, by approximating the vertical direction of the decision boundary in the sample space,
\cite{Moosavidezfooli2015DeepFool} proposes to find the closest adversarial examples to the input with an iterative algorithm.
\cite{moosavi2017universal} further studies the existence of universal perturbations in state-of-the-art deep neural networks.
They suggest the phenomenon is resulted from the high curvature regions on the decision boundary
\cite{moosavi2017analysis}.

These works have built both intuitive and theoretical understanding of the adversarial examples under the Euclidean metric.
However, studying adversarial examples by the Euclidean metric has its limitations.
Intrinsically, for neural networks, the adversarial attacking is about the correlation between the input space and the output space.
Due to the complexity of the networks,
it is hard to explain why small perturbation in the input space can result in large variation in the output space.
Many previous attack methods presume the input space is flat,
thus the gradient with respect to the input gives the fastest changing direction in the output space.
However, if we regard the model output as the likelihood of the discrete distribution,
and regard the model input as the pullback of the output,
a meaningful distance measure for the likelihood will not be linear,
making the sample space a manifold measured by a non-linear Riemannian metric.
This motivates us to adopt the {\bf{Fisher information matrix (FIM)}} of the input as a metric tensor to measure the vulnerability of deep learning models.

The significance of introducing the Fisher information metric is three folds.
First, the FIM is the Hessian matrix of the Kullback-Leibler (KL) divergence,
which is a meaningful metric for probability distributions.
Second, the FIM is symmetrical and positive semi-definite,
making the optimization on the matrix easy and efficient.
Third, the FIM is invariant to reparameterization as long as the likelihood does not change.
This is particularly important for bypassing the influence of irrelevant variables (e.g. different network structures),
and identifying the true cause for the vulnerability of deep learning models.

Based on these insights, we propose a novel algorithm to attack the neural networks.
In our algorithm, the optimization is described by a constrained quadratic form of the FIM,
where the optimal adversarial perturbation is given by the eigenvector,
and the eigenvalues reflect the local vulnerability.
Compared with previous attacking methods,
our algorithm can efficiently characterize multiple adversarial subspaces with the eigenvalues.
In order to overcome the difficulty in computational complexity,
we then introduce some numerical tricks to make the optimization work on large datasets.
We also give a detailed proof for the optimality of the adversarial perturbations under certain technical conditions,
showing that the adversarial perturbations obtained by our method will not be ``compressed'' during the mapping of networks,
which has contributed to the vulnerability of deep learning models.

Furthermore, we perform binary search for the least adversarial perturbation that can fool the networks,
so as to verify the eigenvalues' ability to characterize the local vulnerability:
the larger the eigenvalues are, the more vulnerable the model is to be attacked by the perturbation of corresponding eigenvectors.
Hence we adopt the eigenvalues of the FIM as features,
and train an auxiliary classifier to detect the adversarial attacks with the eigenvalues.
We perform extensive empirical evaluations,
demonstrating that the eigenvalues are of good distinguishability for defending against many state-of-the-art attacks.

Our main contributions in this paper are summarized as follows:
\begin{itemize}
  \item{We propose a novel algorithm to attack deep neural networks based on information geometry.
       The algorithm can characterize multiple adversarial subspaces in the neighborhood of a given sample,
       and achieves high fooling ratio under various conditions.}
  \item{We propose to adopt the eigenvalues of the FIM as features to detect the adversarial attacks.
        Our analysis shows the classifiers with the eigenvalues being their features are robust to various state-of-the-art attacks.}
  \item{We provide a novel geometrical interpretation for the deep learning vulnerability.
        The theoretical results confirm the optimality of our attack method,
        and serve as a basis for characterizing the vulnerability of deep learning models.}
\end{itemize}

\section{Preliminaries}
\paragraph{Fisher information}
The Fisher information is initially proposed to measure the variance of the likelihood estimation given by a statistical model.
Then the idea was extended by introducing differential geometry to statistics \cite{amari2007methods}.
By considering the FIM of the exponential family distributions as the Riemannian metric tensor,
Chenstov further proves that the FIM as a Riemannian measure is the only invariant measure for distributions.
Specifically, let $p(x|\bm{z})$ be a likelihood function given by a statistical model,
where $\bm{z}$ is the model parameter, the Fisher information of $\bm{z}$ has the following equivalent forms:
\begin{align}
  \bm{G}_{\bm{z}}&=\mathbb{E}_{x|\bm{z}}[(\nabla_{\bm{z}}\log{p(x|\bm{z})})(\nabla_{\bm{z}}\log{p(x|\bm{z})})^{T}]\nonumber{}\\
  &=\mathbb{D}_{x|\bm{z}}[\nabla_{\bm{z}}\log{p(x|\bm{z})}]\nonumber{}\\
  &=-\mathbb{E}_{x|\bm{z}}[\nabla^{2}_{\bm{z}}\log{p(x|\bm{z})}],
\end{align}
where $\mathbb{D}_{x|\bm{z}}[.]$ denotes the variance under distribution $p(x|\bm{z})$.

When the FIM is adopted as a Riemannian metric tensor, it enables a connection between statistics and differential geometry.
It is proved that the manifold composed of exponential family distributions is flat under the e-connection,
and the manifold of mixture distributions is flat under the m-connection \cite{amari2007methods}.
The significance is that the metric only depends on the distribution of the model output,
i.e., the FIM is invariant to model reparameterization, as long as the distribution is not changed.
For example, \cite{amari1998natural} shows the steepest direction in the statistical manifold is given by the natural gradient,
which is invariant to reparameterization and saturation-free.

\paragraph{Adversarial attacks}
Many methods are proposed to generate adversarial examples.
The fast gradient method (FGM) and the one-step target class method (OTCM) are
two basic methods that simply adopt the gradient w.r.t. the input as the adversarial perturbation \cite{kurakin2016adversarial}.
The basic iterative method (BIM) performs an iterative FGM update for the input samples with less modifications \cite{Kurakin2016Physical},
which is a more powerful generalization of the ones-step attacks.
Several attack strategies, including the optimization based attack \cite{Liu2016Delving} and the C\&W attack \cite{Carlini2016Towards},
are proposed to craft the adversarial examples via optimization.
The adversarial examples of C\&W attack are proved to be highly transferable between different models,
and can almost completely defeat the defensive distillation mechanism \cite{Papernot2015distillation}.

\paragraph{Adversarial defenses}
The defense against the adversarial examples can be generally divided into the following categories.
The adversarial training takes the adversarial examples as part of the training data,
so as to regularize the models and enhance the robustness \cite{Miyato2015vat,Sinha2018certifiable}.
\cite{Katz2017Reluplex} proposes to verify the model robustness based on the satisfiability modulo theory.
The adversarial detecting approaches add an auxiliary classifier to distinguish the adversarial examples \cite{Metzen2017detecting}.
Many detection measurements, including kernel density estimation, Bayesian uncertainty \cite{Feinman2017detect},
Jensen Shannon divergence \cite{Meng2017MagNet},
local intrinsic dimensionality \cite{Ma2018LID},
have been introduced to detect the existence of adversarial attacks.
Despite the success of the above defenses in detecting many attacks,
\cite{Carlini2017counter,Carlini2017magnet} suggest these mechanisms can be bypassed with some modifications of the objective functions.

\section{The adversarial attack under the Fisher information metric}
\begin{figure*}[htbp]
  \setlength{\abovecaptionskip}{0.cm}
  \setlength{\belowcaptionskip}{-0.cm}
  \centering
  \subfigure[MNIST]{\includegraphics[width=0.32\textwidth]{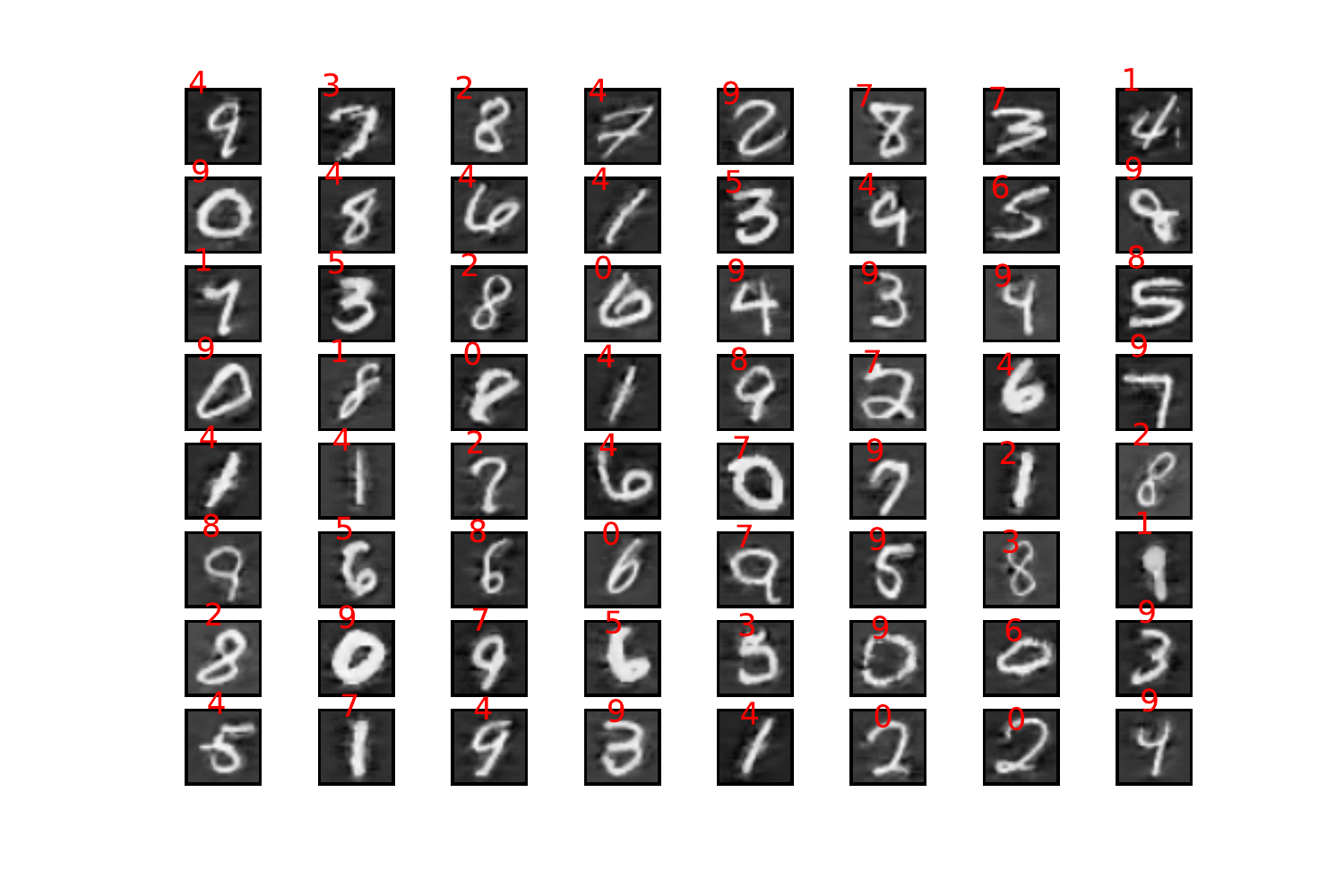}}
  \subfigure[CIFAR-10]{\includegraphics[width=0.32\textwidth]{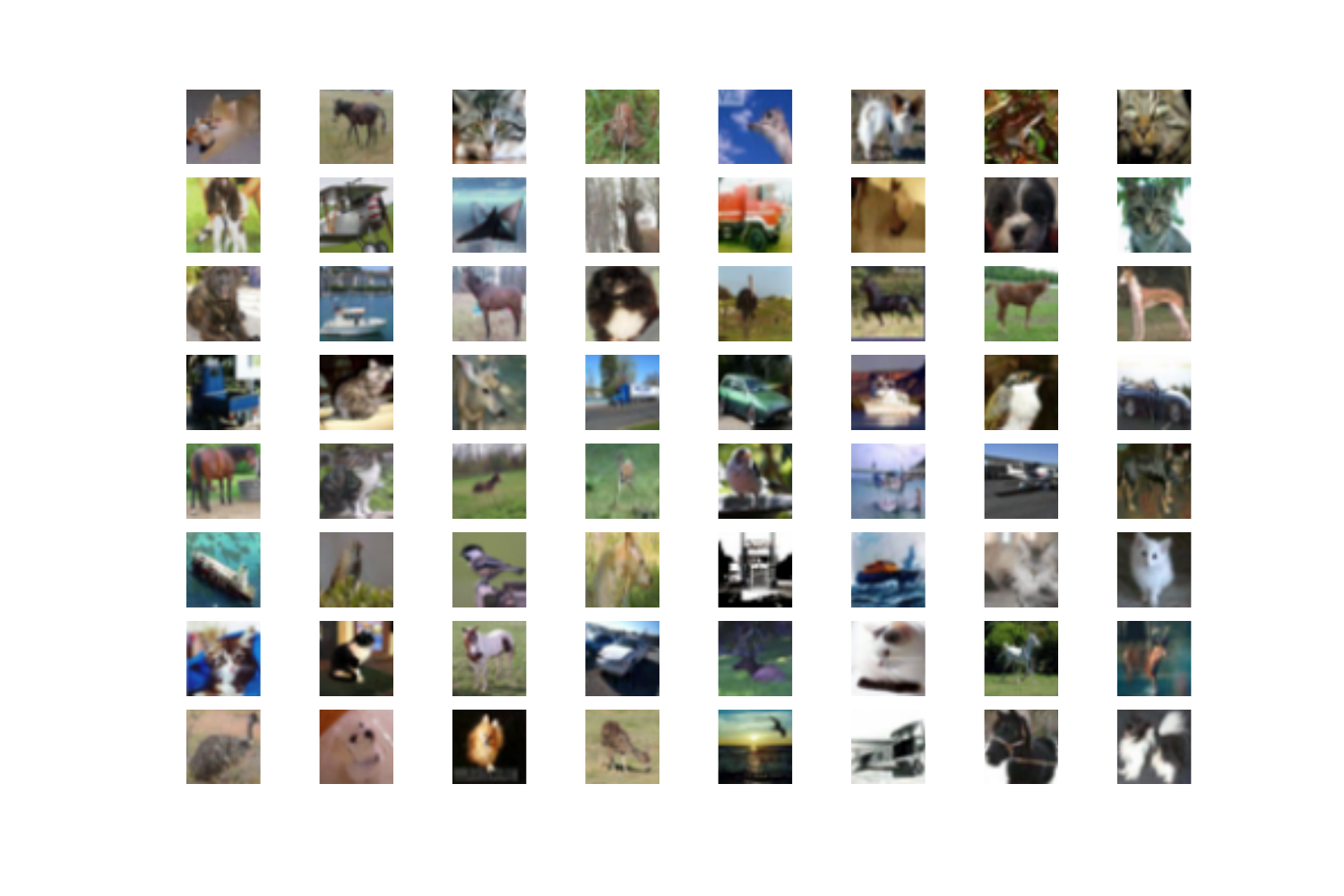}}
  \subfigure[ILSVRC-2012]{\includegraphics[width=0.32\textwidth]{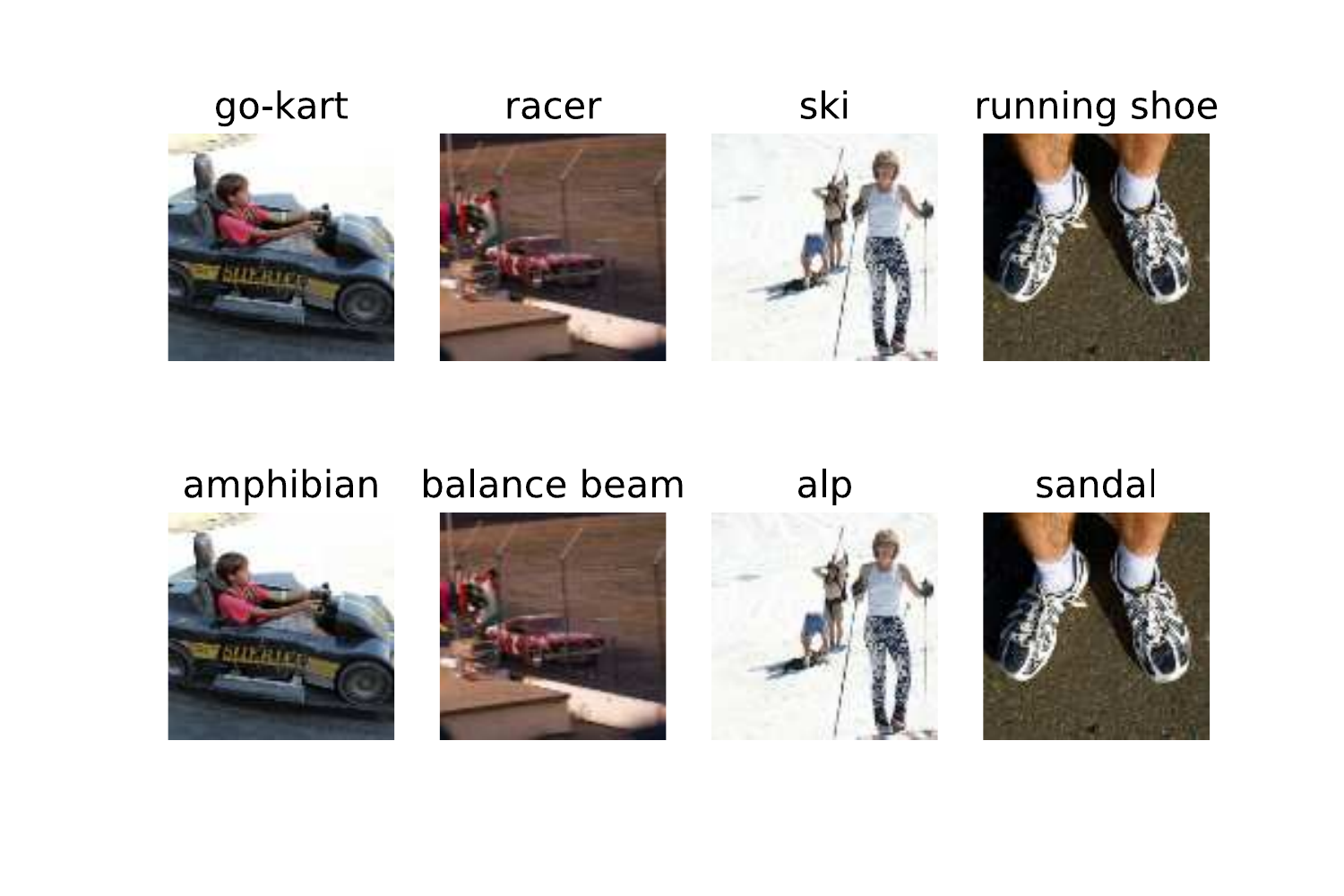}}
  \caption{Visualization for the adversarial examples crafted with our method (Best viewed with zoom-in).
           All the adversarial examples are obtained via one-step update for the original images.
           (a) The model prediction is marked in red numbers.
           (b) All the images here can successfully fool a 14-layer network trained on CIFAR-10.
           (c) The top row shows the original samples, while the second row is the adversarial examples.
           The model prediction is labeled in the top of the images.}
  \label{fig:cifar}
\end{figure*}

\begin{table*}[htbp]
    \centering
    \caption{The comparison for the computation time of OSSA using different numerical methods}
    \label{tab:time}
    \begin{tabular}{ccccccc}
    \toprule
    time (seconds)        & Eigen-decomposition  & Lanczos            & Alias+Lanczos     & Power iteration      & Alias+Power iteration   \\
    \midrule
    CIFAR-10   & $1.49\pm{0.12}$      & $0.30\pm{0.02}$    & $0.15\pm{0.04}$   & $0.28\pm{0.03}$      & $0.25\pm{0.01}$    \\
    ILSVRC-2012           & intractable          & $58.63\pm{3.50}$   & $7.23\pm{0.12}$   & $47.79\pm{3.15}$     & $3.15\pm{0.62}$  \\
    \bottomrule
    \end{tabular}
\end{table*}

\subsection{Proposed algorithm}
In this section, we formalize the optimization of the adversarial perturbations as a constrained quadratic form of the FIM.
As mentioned in the previous section, for classification tasks,
the output of the network can be considered as the likelihood of a discrete distribution.
In information theory, a meaningful metric for different probability distributions is not linear.
Therefore, we start by using the KL divergence to measure the variation of the likelihood distributions.

Consider a deep neural network with its likelihood distribution denoted as $p(y|\bm{x};\bm{\theta})$,
where $\bm{x}$ is the input sample, and $\bm{\theta}$ is the model weights.
Since the model weights are fixed after training, and $\bm{x}$ is the only changeable parameter when attacking,
we omit the model parameters $\bm{\theta}$ in the conditional distribution, and regard $\bm{x}$ as the model parameter.
What the attackers are likely to do is to find a subtle perturbation $\bm{\eta}$,
such that the probability $p(y|\bm{x}+\bm{\eta})$ varies from the correct to the wrong output.
Hence we adopt the KL divergence to measure the variation of the probability $p(y|\bm{x})$.
The optimization objective can be formulated as follows:
\begin{equation}
  \max_{\bm{\eta}}D_{KL}(p(y|\bm{x})||p(y|\bm{x}+\bm{\eta})) \qquad{} \mathrm{s.t.}\ \|\bm{\eta}\|_{2}^{2}=\epsilon,
\end{equation}
where $\epsilon$ is a small parameter to limit the size of the perturbation under the Euclidean metric.
Previous literature has shown that the adversarial examples generally exist in large and continuous regions \cite{Goodfellow2014Explaining}.
such that the models can always be fooled with small perturbation.
Let us assume the perturbation $\|\bm{\eta}\|$ is sufficiently small,
such that the log-likelihood $\log{}p(y|\bm{x}+\bm{\eta})$ can be decomposed using the second-order Taylor expansion.
This yields a simple quadratic form of the FIM:
\begin{align}
  D_{KL}(p(y|\bm{x})\|p(y|\bm{x}+\bm{\eta}))&=\mathbb{E}_{y|\bm{x}}[\log\frac{{p(y|\bm{x})}}{p(y|\bm{x}+\bm{\eta})}]\nonumber{}\\
  &\approx\frac{1}{2}\bm{\eta}^{T}\bm{G}_{\bm{x}}\bm{\eta},
\end{align}
where $\bm{G}_{\bm{x}}=\mathbb{E}_{y|\bm{x}}[(\nabla_{\bm{x}}\log{p(y|\bm{x})})(\nabla_{\bm{x}}\log{p(y|\bm{x})})^{T}]$
is the Fisher information of $\bm{x}$.
Note that the FIM here is not the same as that in \cite{Miyato2015vat}.
Since the expectation is over the observed empirical distribution $p(y|\bm{x})$,
let $p_{i}$ be the probability of $p(y|\bm{x})$ when $y$ takes the $i$-th class,
and let $\mathcal{J}(y,\bm{x})=-\log{p(y|\bm{x})}$ be the loss function of the network,
the matrix can be explicitly calculated by
\begin{equation}
  \bm{G}_{\bm{x}}=\sum_{i}p_{i}[\nabla_{\bm{x}}\mathcal{J}(y_{i},\bm{x})][\nabla_{\bm{x}}\mathcal{J}(y_{i},\bm{x})]^{T}.
\end{equation}

Hence we have a variant form of the objective function, which is given by:
\begin{equation}
\label{objective}
  \max_{\bm{\eta}}\bm{\eta}^{T}\bm{G}_{\bm{x}}\bm{\eta} \quad{}
  \mathrm{s.t.}\ \|\bm{\eta}\|_{2}^{2}=\epsilon,
  \ \mathcal{J}(y,\bm{x}+\bm{\eta})>\mathcal{J}(y,\bm{x}).
\end{equation}

Setting the derivative of the Lagrangian w.r.t. $\bm{\eta}$ to $0$ yields $\bm{G}_{\bm{x}}\bm{\eta}=\lambda\bm{\eta}$.
In general, the optimization can be solved by applying eigen-decomposition for $\bm{G}_{\bm{x}}$,
and assigning the eigenvector with the greatest eigenvalue to $\bm{\eta}$.
Note that eigenvector gives a straight line, not a direction,
i.e., multiplying $\bm{\eta}$ by $-1$ does not change the value of the quadratic form.
Therefore, we add an additional constraint $\mathcal{J}(y,\bm{x}+\bm{\eta})>\mathcal{J}(y,\bm{x})$ here,
guaranteeing that the adversarial examples obtained by $\bm{x}+\bm{\eta}$ will always attain higher loss than the normal samples.

The significance of our method is as follows.
If we consider $D_{KL}(p(y|\bm{x})||p(y|\bm{x}+\bm{\eta}))$ as a function of $\bm{\eta}$,
the Fisher information is exactly the Hessian matrix of the infinitesimal KL divergence.
This implies that the vulnerability of deep learning models can be described by the principal curvature of KL divergence.
Therefore, given an input sample $\bm{x}$,
the eigenvalues of the FIM represent the robustness in the subspaces of corresponding eigenvectors.
The larger the eigenvalues are,
the more vulnerable the model is to be attacked by the adversarial perturbations in the subspaces of corresponding eigenvectors.
This allows us to efficiently characterize the local robustness using the eigenvalues of the FIM.

\subsection{Optimization strategies}
As mentioned before,
the simplest approach to solve the objective function (\ref{objective}) is to calculate the greatest eigenvector of $\bm{G}_{\bm{x}}$.
However, such optimization can be impractical for large datasets.
One main obstacle is that $\bm{G}_{\bm{x}}$ is computed explicitly.
When the image size is large, the exact eigen-decomposition of $\bm{G}_{\bm{x}}$ becomes inefficient and memory consuming.
In order to reduce the computational complexity,
the critical part is to avoid the access to the explicit form of $\bm{G}_{\bm{x}}$.
This can be achieved by computing $\bm{G}_{\bm{x}}\bm{\eta}$ alternatively.
Let $\bm{g}_{y}=\nabla_{\bm{x}}\mathcal{J}(y,\bm{x})$ be the gradient of the class $y$ loss w.r.t. the input $\bm{x}$.
Since the FIM has the form $\bm{G}_{\bm{x}}=\mathbb{E}_{y|\bm{x}}[\bm{g}_{y}\bm{g}_{y}^{T}]$,
by putting $\bm{\eta}$ into the expectation we obtain
\begin{equation}
  \bm{G}_{\bm{x}}\bm{\eta}=\mathbb{E}_{y|\bm{x}}[(\bm{g}_{y}^{T}\bm{\eta})\bm{g}_{y}].
\end{equation}
This allows us to calculate the inner product first, so as to avoid dealing with $\bm{G}_{\bm{x}}$ explicitly.
After $\bm{\eta}$ converges, the greatest eigenvalue has the form $\mathbb{E}_{y|\bm{x}}[(\bm{g}_{y}^{T}\bm{\eta})^{2}]$.

Specifically, when computing the greatest eigenvector of $\bm{G}_{\bm{x}}$,
a naive approach with the power iteration can be adopted to accelerate the eigen-decomposition.
In Step $k$, the power iteration is described by the recurrence equation $\bm{\eta}_{k+1}=\frac{\bm{G}_{\bm{x}}\bm{\eta}_{k}}{\|\bm{G}_{\bm{x}}\bm{\eta}_{k}\|}$.
The iteration thus becomes
\begin{equation}
  \bm{\eta}_{k+1}=\frac{ \mathbb{E}_{y|\bm{x}}[(\bm{g}_{y}^{T}\bm{\eta}_{k})\bm{g}_{y}] }{ \|\mathbb{E}_{y|\bm{x}}[(\bm{g}_{y}^{T}\bm{\eta}_{k})\bm{g}_{y}]\| }.
\end{equation}

Similar approach can be adopted when computing the top $m$ eigenvalues and eigenvectors.
The Lanczos algorithm, which also does not require the direct access to $\bm{G}_{\bm{x}}$,
is an efficient eigen-decomposition algorithm for Hermitian matrices \cite{Calvetti1994Lanczos}.
The algorithm is particularly fast for sparse matrices.
Since $\bm{G}_{\bm{x}}$ is the pullback of a lower dimensional probability space,
this guarantees the efficiency of our implementation.

Additionally, the expectation term in the exact computation of $\bm{G}_{\bm{x}}$ requires to sum over the support of $p(y|\bm{x})$,
which is still inefficient for the datasets with large number of categories. In practice,
the estimation of the integral can be simplified by the Monte Carlo sampling from $p(y|\bm{x})$ with less iterations.
The sampling iterations are set to be approximately $1/5$ number of the classes.
Despite the simplicity, we empirically find the effectiveness is not degraded by the Monte Carlo approximation.
The randomized sampling is performed using the alias method \cite{Marsaglia2004Alias},
which can efficiently sample from high dimensional discrete distribution with $O(1)$ time complexity.

\begin{algorithm}[htbp]
\label{algo1}
  \LinesNumbered
            \caption{One Step Spectral Attack (OSSA)\\Implemented with power iteration+alias sampling}
            \KwIn{input sample $\bm{x}$, corresponding labels $y$,
                  a deep learning model with the output $p(y|\bm{x})$ and the loss $\mathcal{J}(y,\bm{x})$.}
            \KwOut{the perturbation $\bm{\eta}$, the greatest eigenvalue $\lambda^{*}$.}

              Initialize $\bm{\eta}$ as an random vector with unit norm\;
              Initialize the alias table with $p(y|\bm{x})$\;
              \While{$\bm{\eta}$ not converged}
              {
                Update $\bm{\eta}\leftarrow \mathbb{E}_{y|\bm{x}}[(\bm{g}_{y}^{T}\bm{\eta})\bm{g}_{y}]$ using alias sampling\;
                Normalize $\bm{\eta}\leftarrow \frac{\bm{\eta}}{\|\bm{\eta}\|_{2}}$\;
              }
              The greatest eigenvalue $\lambda^{*} \leftarrow \mathbb{E}_{y|\bm{x}}[(\bm{g}_{y}^{T}\bm{\eta})^{2}]$\;
              \If{$\mathcal{J}(\bm{x}+\bm{\eta})\leq{\mathcal{J}(\bm{x})}$}
              {
                $\bm{\eta}\leftarrow{-\bm{\eta}}$\;
              }
\end{algorithm}
In our experiments, we only use the randomization trick for ILSVRC-2012.
Table \ref{tab:time} shows the comparison for the time consumption of the aforementioned methods.
To summarize, the algorithm procedure of the alias method+power iteration implementation is shown in Algorithm \ref{algo1}.
In Figure \ref{fig:cifar}, we also illustrate some visualization of the adversarial examples crafted with our method.

\subsection{Geometrical interpretation}

Characterizing the vulnerability of neural networks is an important question for studying adversarial examples.
Under the Euclidean metric, \cite{Sinha2018certifiable} has suggested that identifying the worst case perturbation in ReLU networks is NP-hard.
In this subsection, we give an explanation for the vulnerability of deep learning from a different aspect.
Our aim is to prove that under the Fisher information metric,
the perturbation obtained by our algorithm will not be ``compressed'' through the mapping of networks,
which has contributed to the vulnerability of deep learning.

Geometrically, let $\bm{z}=f(\bm{x})$ be the mapping through the neural network,
where $\bm{z}\in{(0,1)^{k}}$ is the continuous output vector of the softmax layer,
with $p(y|\bm{z})=\prod_{i}\bm{z}_{i}^{y_{i}}$ being a discrete distribution.
We can conclude the FIM
$\bm{G}_{\bm{z}}=\mathbb{E}_{y|\bm{z}}[(\nabla_{\bm{z}}\log{p(y|\bm{z})})(\nabla_{\bm{z}}\log{p(y|\bm{z})})^{T}]$
is a non-singular diagonal matrix.
The aforementioned Fisher information $\bm{G}_{\bm{x}}$ is thus interpreted as a Riemannian metric tensor induced from $\bm{G}_{\bm{z}}$.
The corresponding relationship is
\begin{equation}
  \bm{\eta}^{T}\bm{G}_{\bm{x}}\bm{\eta}=\bm{\eta}^{T}\bm{J}_{f}^{T}\bm{G}_{\bm{z}}\bm{J}_{f}\bm{\eta},
\end{equation}
where $\bm{J}_{f}$ is the Jacobian of $\bm{z}$ w.r.t. $\bm{x}$.
Note that for most neural networks, the dimensionality of $\bm{z}$ is much less than that of $\bm{x}$.
making $\bm{J}_{f}\bm{\eta}$ a mapping for $\bm{\eta}$ from high dimensional data space to low dimensional probability space.
This means $f$ is a surjective mapping and $\bm{G}_{\bm{x}}$ is a degenerative metric tensor.
Therefore, the geodesic distance in the probability space is always no larger than the corresponding distance in the data space.
Using the inequality, we can define the concept of optimal adversarial perturbation, formulated as follows.

\newtheorem{def1}{Definition}
\begin{def1}
  Let $\mathcal{N}$ and $\mathcal{M}$ be two Riemannian manifolds with the FIMs $\bm{G}_{\bm{z}}$ and $\bm{G}_{\bm{x}}$ being their metric tensor respectively.
  Let $f:\mathcal{M}\rightarrow{\mathcal{N}}$ be the mapping of the neural network. For $\bm{x}\in{\mathcal{M}}$, an adversarial perturbation $\bm{\eta}\in{T_{\bm{x}}\mathcal{M}}$ is {\bf{optimal}} if $f(\bm{x})$ is an isometry for the geodesic determined by the exponential mapping $\mathrm{Exp}_{\bm{x}}(\bm{\eta}):T_{\bm{x}}\mathcal{M}\rightarrow{\mathcal{M}}$.
\end{def1}

\begin{def1}
 Let $f:\mathcal{M}\rightarrow{\mathcal{N}}$ be a smooth mapping,
 and $f_{*}:T_{\bm{x}}\mathcal{M}\rightarrow{T_{f(\bm{x})}}\mathcal{N}$ be the derivative of $f$.
 The mapping $f$ is a {\bf{submersion}} if $f$ is surjective and $f_{*}$ is surjective,
 and a {\bf{Riemannian submersion}} if $f_{*}$ is an isometry on the horizontal bundle $H=ker\ f_{*}^{\bot}$.
\end{def1}

The definitions show the optimal perturbations span only on the horizontal bundles.
Thus the conclusion is as follows.

\newtheorem{thr1}{Theorem}
\begin{thr1}
\label{theorem1}
  Let $\bm{J}_{f}$ be the Jacobian field of $f(\bm{x})$ w.r.t. $\bm{x}$.
  If $\bm{J}_{f}\bm{J}_{f}^{T}$ is non-singular,
  and $f:\mathcal{M}\rightarrow{\mathcal{N}}$ is a smooth mapping,
  then a sufficiently small perturbation $\bm{\eta}\in{T_{\bm{x}}\mathcal{M}}$ obtained by Algorithm \ref{algo1} is optimal.
\end{thr1}

\begin{proof}
    For the neural network $f$, we define $V_{\bm{x}} \subset T_{\bm{x}}\mathcal{M}$,
    the vertical subspace at a point $\bm{x}\in{\mathcal{M}}$, as the kernel of the FIM $\bm{G}_{\bm{x}}$.
    In our algorithm, we always apply the greatest eigenvector in the FIM $\bm{G}_{\bm{x}}$ as the adversarial perturbation.
    Given a smooth network $f$, because $\bm{J}_{f}\bm{J}_{f}^{T}$ is always non-singular,
    the first eigenvalue in the FIM is always larger than zero,
    which corresponds to the non-degenerative direction.
    Therefore the adversarial perturbation $\bm{\eta}\in{T_{\bm{x}}}\mathcal{M}$ obtained is always in the horizontal bundle,
    i.e., $\bm{\eta}\in{H}$.
    By definition, $f_{*}:T_{\bm{x}}\mathcal{M}\rightarrow{T_{f(\bm{x})}}\mathcal{N}$ will be an isometry for the horizontal sub-bundles.
    Then $f$ will also be an isometry for the geodesic determined by $\mathrm{Exp}_{\bm{x}}(\bm{\eta}):T_{\bm{x}}\mathcal{M}\rightarrow{\mathcal{M}}$.
\end{proof}

In a broader sense, the theorem confirms the validity of our proposed approach,
and serves as a basis for characterizing the vulnerability of deep learning models.
Note that the theorem is concluded without any assumption for the network structures.
The optimality can thus be interpreted as a generalization of the excessive linearity explanation \cite{Goodfellow2014Explaining}.
The statement shows that the linearity may not be a sufficient condition for the vulnerability of neural networks.
Using our algorithm, similar phenomenon can be reproduced in a network with smooth activations, e.g. the exponential linear unit \cite{clevert2015elu}.

\subsection{Experimental evaluation}

In this section, by presenting experimental evaluations for the properties of the adversarial attacks,
we show the ability of our attack method to fool deep learning models, and characterize the adversarial subspaces.
The experiments are performed on three standard benchmark datasets MNIST,
CIFAR-10 \cite{krizhevsky2009learning}, and ILSVRC-2012 \cite{ILSVRC2012}.
The pixel values in the images are constrained in the interval [0.0, 1.0].
We adopt three different networks for the three datasets respectively: LeNet-5, VGG, and ResNet-152 \cite{resnet2015He}.
The VGG network adopted here is a shallow variant of the VGG-16 network \cite{simonyan2014very},
where the layers from conv4-1 to conv5-3 are removed to reduce redundancy.
We use the pre-trained ResNet-152 model integrated in TensorFlow.
In our experiments, all the adversarial perturbations are evaluated with $\ell_{2}$ norm.

\paragraph{White-box attack}
In the first experiment, we perform comparisons for the ability of our method to fool the deep learning models.
The comparison is made between two one-step attack methods, namely FGM and OTCM, and their iterative variants.
The target class in OTCM is randomly chosen from the set of incorrect classes.
Similar to the relationship between FGM and BIM, by computing the first eigenvector of FIM in each step,
it is a natural idea to perform our attack strategy iteratively.
For the iterative attack, we set the perturbation size $\epsilon=0.05$.
We only use the samples in the test set (validation set for ILSVRC-2012) to craft the adversarial examples.
\begin{figure}[h]
  \centering
  \subfigure[One-step attack]{\includegraphics[width=0.234\textwidth]{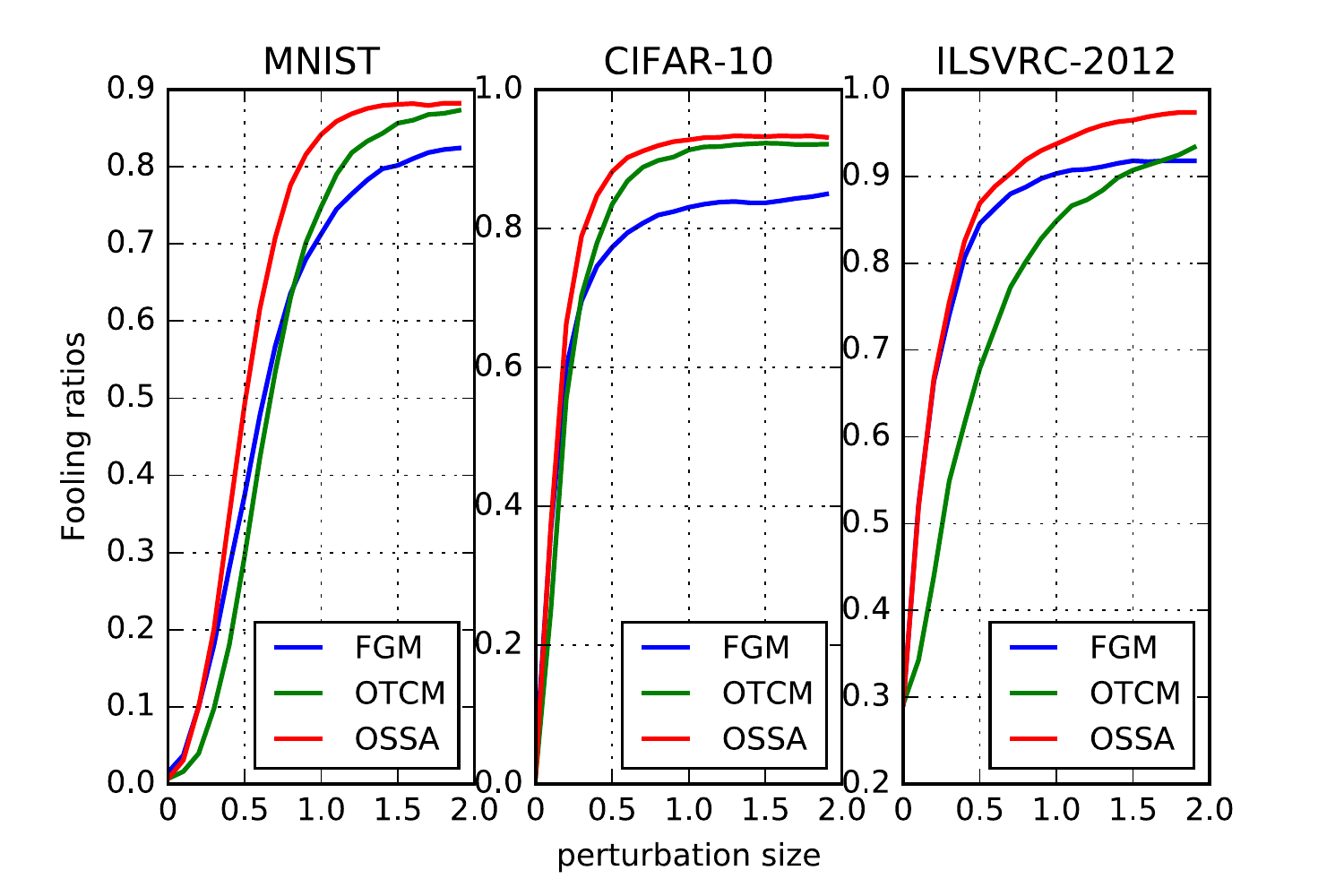}}
  \subfigure[Iterative attack]{\includegraphics[width=0.234\textwidth]{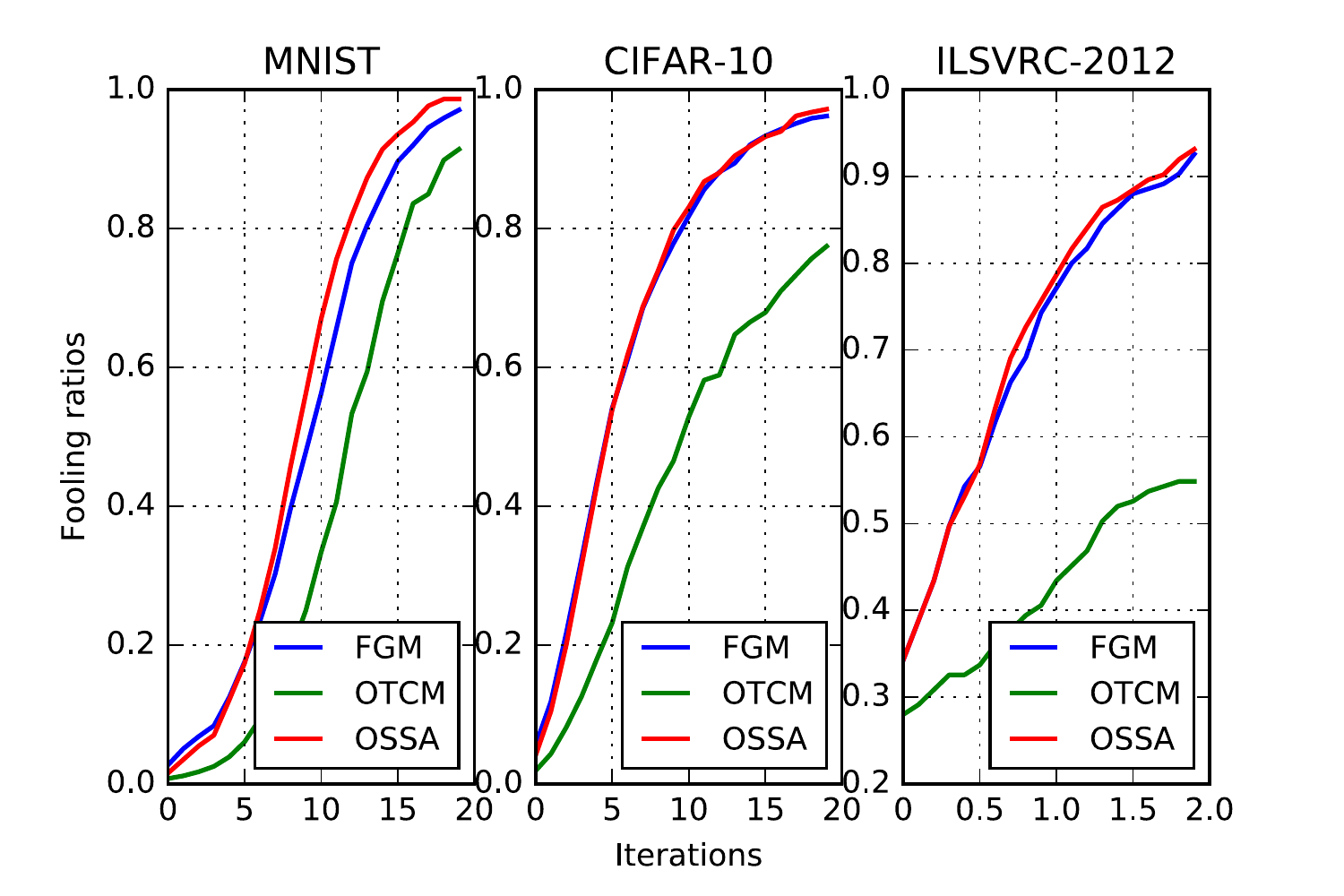}}
  \caption{(a) The models' misclassification rates increase with the perturbation size.
  (b) The models' misclassification rates increase with the number of iterations.
    When performing the iterative attacks, we set perturbation size to 0.05, 0.025, 0.0125 for the three datasets respectively.}
  \label{fig_plot}
\end{figure}

The results are illustrated in Figure \ref{fig_plot}.
Observe that our proposed method quickly attains a high fooling rate with smaller values of $\|\bm{\eta}\|$.
For one-step attacks, our method achieves 90\% fooling ratio with the $\ell_{2}$ perturbation norms 2.1, 1.4, and 0.7
on MNIST, CIFAR-10, and ILSVRC-2012 respectively.
This implies that the eigenvectors as adversarial perturbations is a better characterization for model robustness than the gradients.
Another evidence is shown in Table \ref{tab:fool_cmp},
where we conduct binary search to find the mean of the least perturbation norm on three datasets.
Our approach achieves higher fooling ratios than the gradient-based method,
with a smaller mean of the least perturbation norm.
Both of the results are consistent with our conclusion in previous sections.
\begin{table}[]
\caption{The fooling rates and the mean of least $\ell_{2}$ perturbation norms under two one-step attack strategies}
\label{tab:fool_cmp}
\begin{tabular}{@{}ccccccc@{}}
\toprule
                 & \multicolumn{2}{c}{\textbf{MNIST}} & \multicolumn{2}{c}{\textbf{CIFAR}} & \multicolumn{2}{c}{\textbf{ILSVRC}} \\
\textbf{Attacks} & mean           & rate \%           & mean           & rate \%           & mean            & rate \%           \\ \midrule
FGM              & 2.11           & 94.98             & 1.11           & 95.21             & 0.48            & 100               \\
OSSA             & 1.80           & 95.68             & 1.06           & 97.85             & 0.47            & 100               \\ \bottomrule
\end{tabular}
\end{table}

\paragraph{Black-box attack}
In the real world, the black-box attack is more common than the white-box attacks.
It is thus important to analyze the transferability between different models.
In this experiment, we show the ability of our attack approach to transfer across different models,
particularly the models regularized with adversarial training.
The experiment is performed on MNIST, with four different networks: LeNet, VGG, and their adversarial training variants,
which is referred to as LeNet-adv and VGG-adv here.
For the two variant networks, we replace all the ReLU activations with ELUs,
and train the network with adversarial training using FGM.
All of the above networks achieve more than 99\% accuracy on the test set of MNIST.
To make the comparison fair, we set $\epsilon=2.0$ for all the tested attack methods.

The results of this experiment are shown in Table \ref{transfer_table}.
The cross-model fooling ratios are obviously asymmetric between different models.
Specifically, the adversarial training plays an important role for defending against the attack.
The models without adversarial training produce 22.51\% error rate in average on the the models with adversarial training,
while the reversed case is 80.52\% in average.
Surprisingly, the adversarial examples crafted from the models with adversarial training yield high fooling ratios on the two normal networks.
Whereas a heuristic interpretation is that the perturbations obtained by OSSA correspond to only one subspace,
making the adversarial training less specific for our attack strategy,
the reason of the phenomenon requires further investigation.

\begin{table}[]
\caption{The cross-model fooling ratios on MNIST using OSSA.}
\label{transfer_table}
\begin{tabular}{@{}ccccc@{}}
\toprule
\multicolumn{1}{l}{Fooling rates}   & \multicolumn{4}{c}{\textbf{Crafted from}} \\
\textbf{Tested on}                  & LeNet   & VGG     & LeNet-adv   & VGG-adv      \\ \midrule
LeNet                               & 100.0   & 62.02   & 88.49       & 82.20        \\
VGG                                 & 53.64   & 100.0   & 76.93       & 74.45        \\
LeNet-adv                           & 27.92   & 17.83   & 100.0       & 90.04        \\
VGG-adv                             & 15.06   & 29.24   & 94.15       & 100.0        \\ \bottomrule
\end{tabular}
\end{table}

\paragraph{Characterizing multiple adversarial subspaces}
\begin{figure}[htp]
  \centering
  \subfigure[MNIST]{\includegraphics[width=0.23\textwidth]{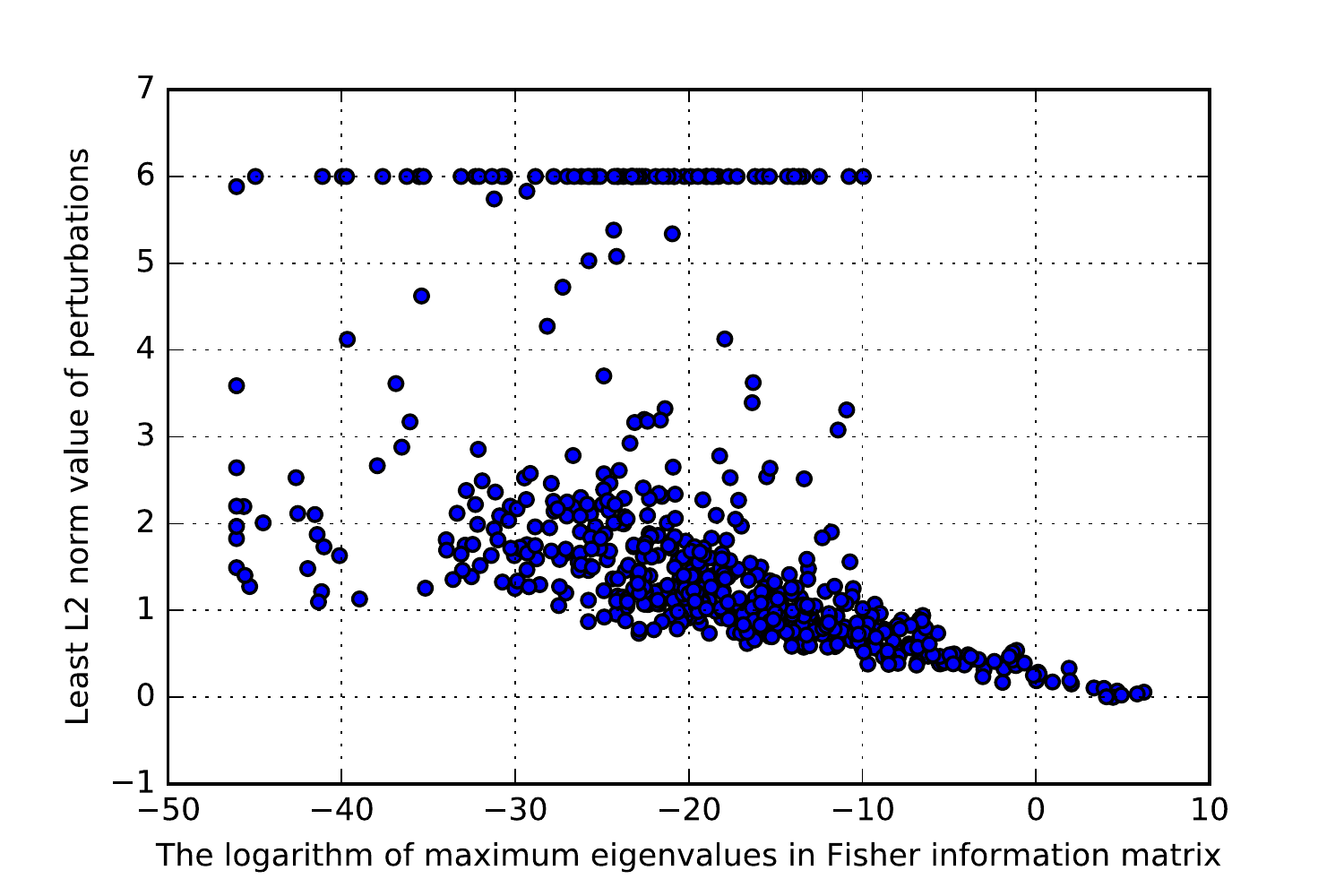}}
  \subfigure[CIFAR-10]{\includegraphics[width=0.23\textwidth]{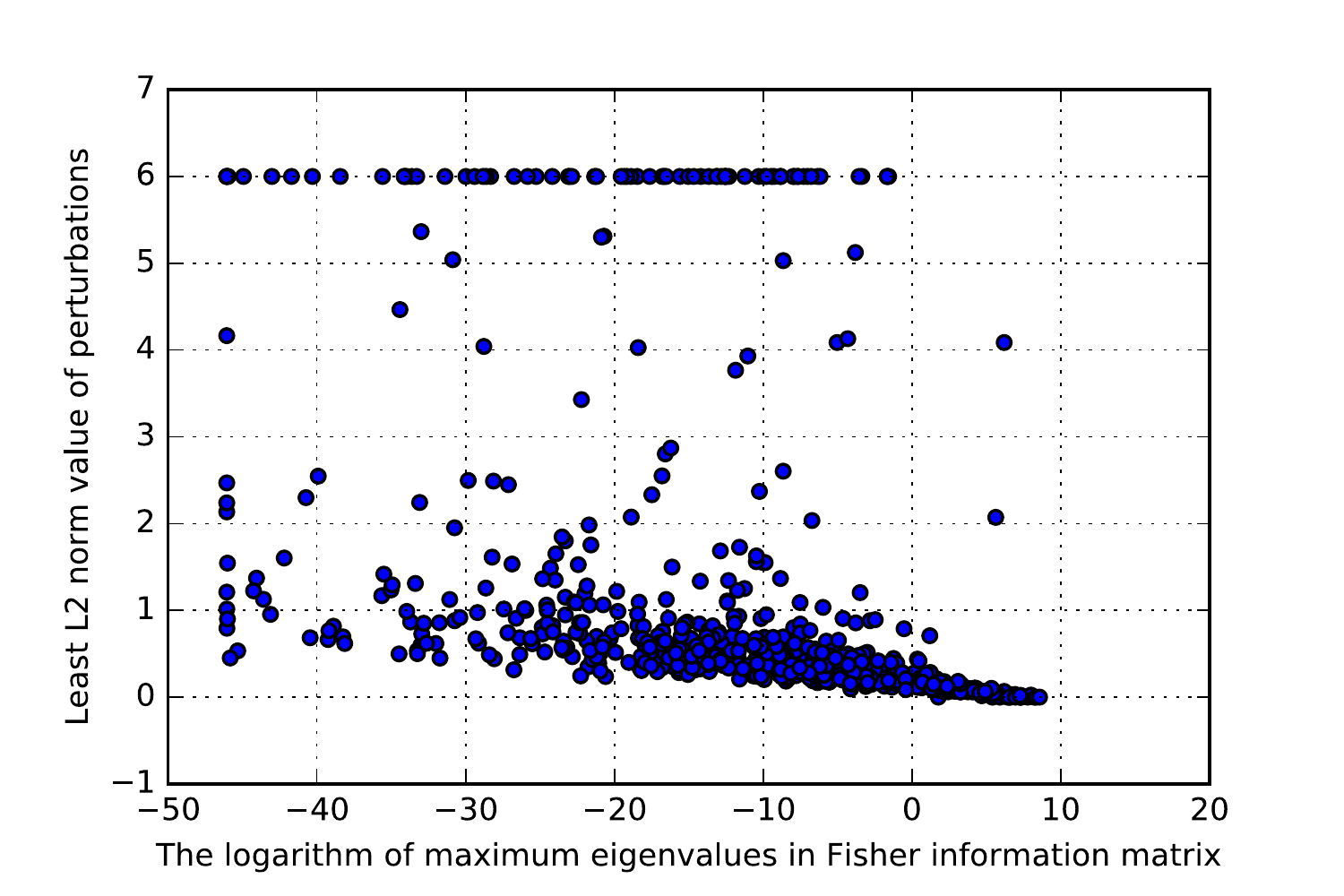}}
  \caption{Using 800 random samples,
           the scatter illustrates the relationship between the least $\ell_{2}$ perturbation norm and the maximum eigenvalues.
           The least $\ell_{2}$ perturbation size is obtained via binary search in the interval [0.0, 6.0].
           The horizontal axis is shown in logarithm.}
  \label{fig_scatter}
\end{figure}
As shown in the previous sections, the eigenvalues in FIM can be applied to measure the local robustness.
We thus perform experiments to verify the correlation between the model robustness and the eigenvalues.
In Figure \ref{fig_scatter}, we show the scatter of 800 randomly selected samples in the validation set of MNIST and CIFAR-10.
The horizontal axis is the logarithm of the eigenvalues, and the vertical axis is the least adversarial perturbation size,
i.e., the least value of $\|\bm{\eta}\|_{2}$ to fool the network.
The value is obtained via binary search between the interval $[0.0, 6.0]$.
Most adversarial examples can successfully fool the model in this range.
The result shows an obvious correlation between the eigenvalues and the model vulnerability:
the least perturbations linearly decrease with the exponential increasing of eigenvalues.

A reasonable interpretation is that the eigenvalues reflect the size of the perturbations under the Fisher information metric.
According to our optimality analysis, large eigenvalues can result in isometrical variation for the output likelihood,
which is more likely to fool the model with less perturbation size.
This property is crucial for our following discussion, the adversarial detection,
where we take advantage of the distinguishability of the eigenvalues to detect the adversarial attacks.

\section{The adversarial detection under the Fisher information metric}

\begin{figure}[h]
  \centering
  \subfigure[Eigenvalues' distributions]{\includegraphics[width=0.23\textwidth]{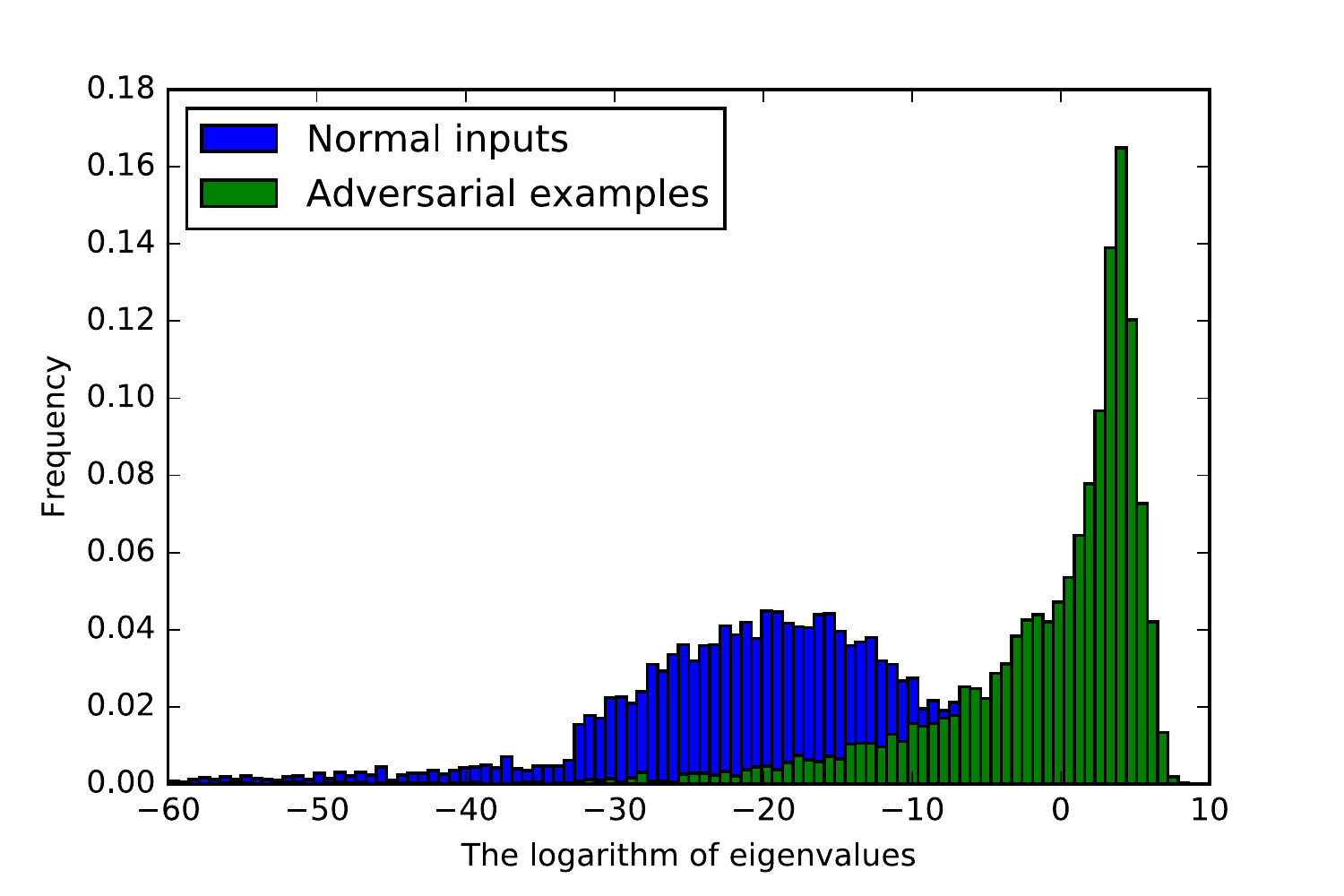}}
  \subfigure[Increment of eigenvalues]{\includegraphics[width=0.23\textwidth]{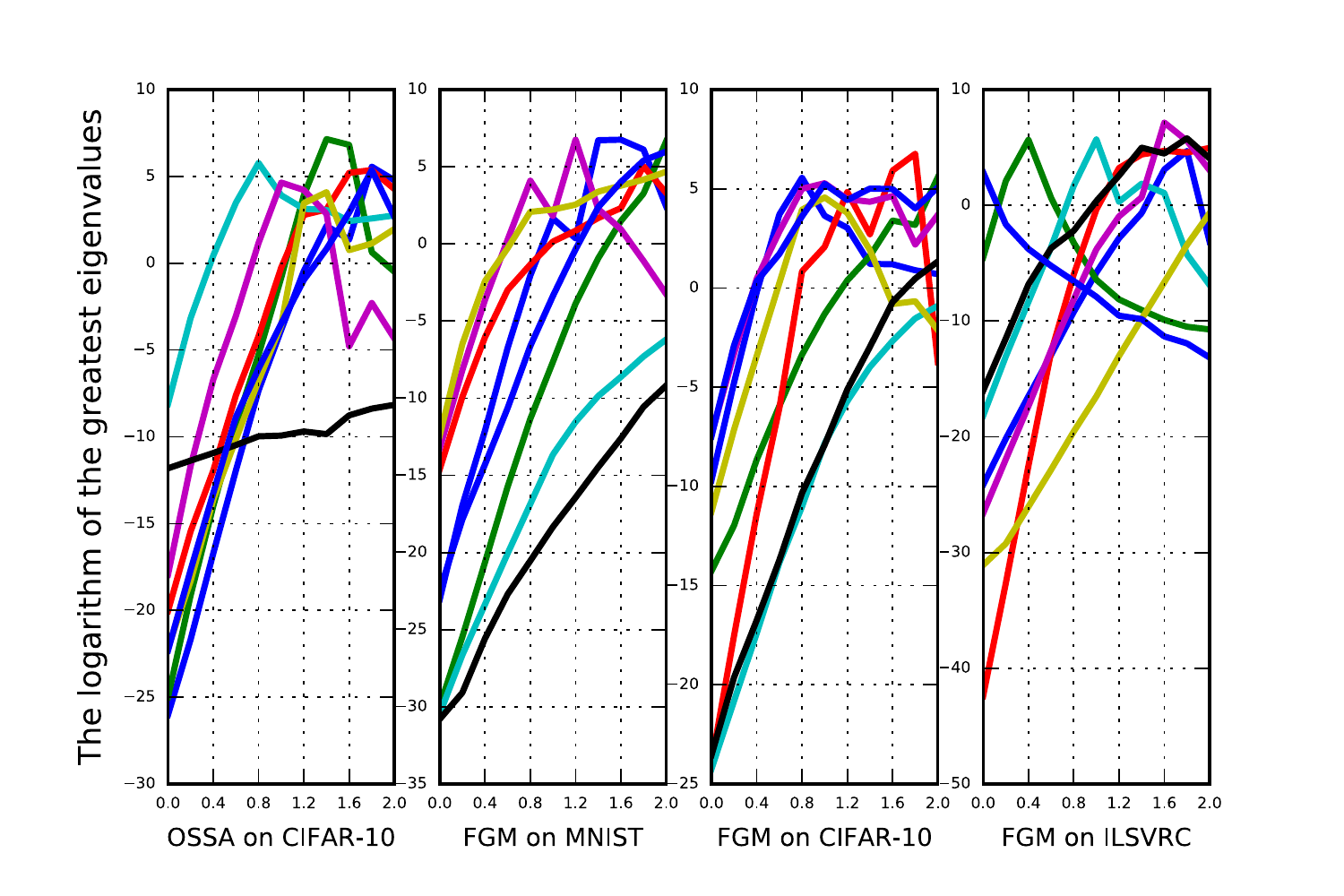}}
  \caption{Some empirical evidence for the distinguishability of the eigenvalues.
  (a) The histograms for the distribution of largest eigenvalues.
  The statistic is performed on all the samples in the test set of MNIST.
  (b) The increment of the eigenvalues along the direction of the adversarial perturbations.
  The samples are randomly sampled from MNIST, CIFAR-10, and ILSVRC-2012.}
  \label{fig_distribution}
\end{figure}

\begin{table*}[]
\centering
\caption{The AUC scores of detecting adversarial attacks using random forest.
         The best are marked with \bf{bold} font.}
\label{AUC_table}
\begin{tabular}{cccccccccccc}
\hline
         & \multicolumn{5}{c}{\textbf{MNIST}}                                                 & \multicolumn{5}{c}{\textbf{CIFAR-10}}                                              \\
AUC (\%) & FGM   & OTCM   & Opt      & BIM      & OSSA$\quad$     & FGM     & OTCM      & Opt      & BIM            & OSSA           \\ \hline
KD       & 78.12          & 95.46    & 95.15    & 98.61    & 84.24$\quad$   & 64.92     & 92.13    & 91.35          & 98.70          & 88.89          \\
BU       & 32.37          & 91.55    & 71.30    & 25.46    & 74.21$\quad$   & 70.40     & 91.93    & 91.39          & 97.32          & 87.44          \\
KD+BU    & 82.43          & 95.78    & 95.35    & 98.81    & 85.97$\quad$   & 76.40     & \textbf{94.45} & 93.77    & 98.90          & 93.54          \\
Ours     & \textbf{96.11} & \textbf{98.47} & \textbf{95.67} & \textbf{99.10} & \textbf{93.13}$\quad$ & \textbf{80.18} & 93.68          & \textbf{99.45} & \textbf{99.43} & \textbf{98.01} \\ \hline
\end{tabular}
\end{table*}

As shown in the previous sections, given an input sample $\bm{x}$,
the eigenvalues in FIM can well describe its local robustness.
In this section, we show how the eigenvalues in FIM can serve as features to detect the adversarial attacks.
Specifically, the detection is achieved by training an auxiliary classifier to recognize the adversarial examples,
with the eigenvalues serving as the features for the detector.
Motivated by \cite{Fawzi2016noise}, besides the normal original samples and the adversarial inputs,
we also craft some noisy samples to augment the detection.
Since the networks are supposed to be robust to some random noise applied to the input,
the set of negative samples should contain both the normal samples and noisy samples,
while the set of positive samples contain the adversarial examples.

In the left of Figure \ref{fig_distribution}, we show a histogram of the eigenvalues distribution.
We adopt the FGM to generate adversarial examples for the samples from MNIST, and evaluate their greatest eigenvalues in FIM.
The histogram shows that the distributions of the eigenvalues for normal samples and adversarial examples are different in magnitude.
The eigenvalues of the latter are densely distributed in larger domain,
while the distribution of the former is approximately an Gaussian distribution with smaller mean.
Although there is overlapping part for the supports of the two distributions,
the separability for the adversarial examples can be largely enhanced by adding more eigenvalues as features.
In the right of Figure \ref{fig_distribution}, using our proposed OSSA,
we illustrate some examples of the eigenvalues increasing along the direction of the adversarial perturbations.
As we predicted, the eigenvalues increase with the increasing of the perturbation size,
showing that the adversarial examples have higher eigenvalues in FIM compared with the normal samples.

The next question is which machine learning classifier should be adopted for the detection.
In out experiments, we empirically find the models are more likely to attain high variance instead of high bias.
The naive Bayes classifier with Gaussian likelihood,
and the random forest classifier yields the best performance among various models.
The success of the former demonstrates that the geometry structure in each subspace is relatively independent.
As for the random forest classifier,
we empirically find that varying the parameters (e.g. the tree depth, the value of $\epsilon$, etc.)
does not significantly affect the AUC scores.
We also find the tree depth not to exceed 5, and more than 20 trees in the random forest yields good performance.
These results imply that our detection with Fisher information enjoys low variance.

In Table \ref{AUC_table}, we adopt the AUC score to evaluate the performance of our random forest classifier under different attacks.
The comparison is made between our approach and two characteristics described in \cite{Feinman2017detect},
namely the kernel density estimation (KD) and the Bayesian uncertainty (BU).
In our experiments, only the top 20 eigenvalues are extracted as the features for classification.
Observe that the detector achieves desirable performance in recognizing the adversarial examples.
The eigenvalues as features outperform KD and BU on both datasets.
In addition, our detector is particularly good at recognizing OSSA adversarial examples.
The AUC scores are 7.16\% and 4.47\% higher than the combination of the other two characteristics.

\begin{table}[htbp]
\caption{The generalization ability for detecting adversarial attacks on MNIST with random forest classifier}
\label{tab:generalization}
\begin{tabular}{@{}cccccc@{}}
\toprule
\multicolumn{1}{l}{AUC (\%)}           & \multicolumn{5}{c}{\textbf{Tested on}} \\
\textbf{Trained on}                    & FGM    & OTCM  & Opt   & BIM   & OSSA  \\ \midrule
FGM                                    & 94.31  & 91.92 & 90.78 & 91.87 & 92.13 \\
OTCM                                   & 98.55  & 98.96 & 98.26 & 97.78 & 98.57 \\
Opt                                    & 95.18  & 95.30 & 96.90 & 97.15 & 96.11 \\
BIM                                    & 98.10  & 96.00 & 97.09 & 98.57 & 96.35 \\
OSSA                                   & 91.17  & 91.47 & 89.77 & 89.47 & 89.67 \\ \bottomrule
\end{tabular}
\end{table}

In the real world, we cannot presume all the attacks strategies are known before we train the detector.
It is thus important for the features to have sufficient generalization ability.
In Table \ref{tab:generalization}, we show the AUC scores of the detector trained on only one type of adversarial examples.
Observe that most of our results exceed 90\% of AUC scores,
indicating the adversarial examples generated by various methods share similar geometric properties under the Fisher information metric.
Interestingly, the detector trained on OSSA obtain the worst generalization ability among all methods.
We regard this is due to the geometrical optimality of our method.
According to our analysis in the previous section, the adversarial examples of OSSA may distribute densely in more limited subspaces,
resulting in less diversity for generalization.

\section{Conclusion}
In this paper, we have studied the adversarial attacks and detections using information geometry,
and proposed a method unifying the adversarial attack and detection.
For the attacks, we show that under the Fisher information metric,
the optimal adversarial perturbation is the isometry between the input space and the output space,
which can be obtained by solving a constrained quadratic form of the FIM.
For the detection, we observe the eigenvalues of FIM can well describe the local vulnerability of a model.
This property allows us to build machine learning classifiers to detect the adversarial attacks with the eigenvalues.
Experimental results have shown promising robustness on the adversarial detection.

Addressing the adversarial attacks issue is typically difficult.
One of the great challenges is the lack of theoretical tools to describe and analyze the deep learning models.
We are confident that the Riemannian geometry is a promising approach to leverage better understanding for the vulnerability of deep learning.

In this paper, we only focus on the classification tasks, where the likelihood of the model is a discrete distribution.
Besides classification, there are many other tasks which can be formulated as statistical problems,
e.g. Gaussian distribution for regression tasks. Therefore,
investigating the adversarial attacks and defenses on other tasks will be an interesting future direction.

\section*{Acknowledgement}
This work is supported by the National Science Foundation of China
(Nos.\,11771276,
11471208,
61731009
 and 61273298), Shanghai Key Laboratory of Multidimensional Information Processing, East China Normal University, China,
and the Science and Technology Commission of Shanghai
Municipality (No. 14DZ2260800).

\fontsize{9.25pt}{10.25pt}
\bibliographystyle{aaai}
\bibliography{AAAI-ZhaoC-5396}

\end{document}